\documentclass[letterpaper, 10pt, conference]{ieeeconf}

\IEEEoverridecommandlockouts

\usepackage{amsmath,amssymb,amsfonts}
\usepackage{graphicx}
\usepackage{textcomp}
\usepackage{xcolor}
\usepackage{placeins}
\usepackage{comment}
\usepackage{pbalance}
\usepackage[style=ieee,maxbibnames=3,minbibnames=1,doi=false,isbn=false,eprint=false]{biblatex}
\AtEveryBibitem{%
  \clearfield{issn}\clearfield{note}%
  \ifboolexpr{ test {\ifentrytype{misc}} or test {\ifentrytype{online}}
            or test {\ifentrytype{manual}} or test {\ifentrytype{softmisc}} }
    {}{\clearfield{url}\clearfield{urldate}}%
}
\def\BibTeX{{\rm B\kern-.05em{\sc i\kern-.025em b}\kern-.08em
    T\kern-.1667em\lower.7ex\hbox{E}\kern-.125emX}}
\begin{document}
\title{MESSPa: A Multi-modal Embodiment Stimulation Shoulder Pad for \\ Preserving the Sense of Agency in a Detachable Robotic Arm}

\author{Vitvasin Vimolmongkolporn$^{1}$, Guillaume Gourmelen$^{1}$, Yukiko Iwasaki, Hiroyasu Iwata%
\thanks{All are with Department of Modern Mechanical Engineering, Waseda University, Tokyo, Japan
        {\tt\small $^{1}$Corresponding authors: vitvasin.au@fuji.waseda.jp, g.guillaume@aoni.waseda.jp}}%
}

\maketitle

\begin{abstract}
Wearable robots such as supernumerary robotic limbs (SRLs) can augment human capability, but their effectiveness depends strongly on the operator's sense of embodiment, comprising the sense of agency (SoA) and the sense of ownership (SoO). When an SRL is detached from the body so that the operator can act in two places at once, both the sense of agency and task performance tend to degrade. This paper presents the Multi-modal Embodiment Stimulation Shoulder Pad (MESSPa), a wearable interface that combines one-to-one motion transfer with spatially distributed pneumatic (balloon) haptic feedback on the shoulder, with the goal of preserving the sense of agency across attached and detached states. We describe the system design, the control mapping, and a controlled experiment using a $2\times2$ mixed design that crosses the control interface between groups (MESSPa, $N=13$, vs.\ keyboard control, $N=12$ from our previous study) and attachment (on-shoulder/detached) within participants, under single- and dual-task loads. Compared with keyboard control in the detached condition, MESSPa significantly reduced task completion time and significantly improved the sense of agency in the detached state, while narrowing the performance gap between the attached and detached conditions. The sense of ownership did not change significantly. These results indicate that combining motion transfer with positional haptic feedback is an effective way to recover agency when a robotic limb is operated remotely.
\end{abstract}

\textit{Keywords:} Haptic interfaces; Wearable robotic systems --- Orthotics and Exoskeletons; Human--robot collaboration; Sense of embodiment.

\section{Introduction}
In recent years, wearable robots have emerged as a transformative technology with the potential to significantly enhance human capability. These devices, which include exoskeletons \cite{Siviy2023_Exoskeleton} and supernumerary robotic limbs (SRLs) \cite{Bright2017SupernumeraryRL, Parietti2016_SRL, Prattichizzo_extrafinger}, are designed to augment human physical abilities in tasks ranging from industrial work to rehabilitation. A critical determinant of their effectiveness is the degree of embodiment the user experiences \cite{HRI_Embodiment}. The sense of embodiment (SoE) \cite{AgencyandOwnership, BLANKE20097_SOE} is commonly decomposed into the sense of agency (SoA), the feeling of controlling the device, and the sense of ownership (SoO), the feeling that the device is part of one's own body \cite{Haggard2012_SenseofAgency, Kondo2018_bodyOwnership}. Together these shape how naturally and effectively a user interacts with a robotic device.

Because SRLs are worn on the body, they are physically constrained to act near the operator, which limits their use for tasks that require presence in two separate locations. To address this, the concept of a \emph{detachable body} has been proposed \cite{Iwasaki2020}: a modular robotic limb that the operator can leave at a remote site and continue to control through a head-mounted display (HMD) and a robotic head \cite{Vimolmongkolporn2022}. Building on this idea, an experiment-assisting system with a local augmented body was developed for remote work during pandemics \cite{Iwasaki2021}, and detachable manipulation has been extended to end-effectors that relocate themselves to reach distant objects \cite{epfl2024detachable}. More broadly, ubiquitous robotics \cite{Ubiquitous_robotics} envisions modular, networked robots that operate seamlessly across everyday environments. Our previous study \cite{Vimolmongkolporn2023} examined how operators perform with such a modular body in attached versus detached configurations and identified the spatial arrangement between the head and arm that best supports remote operation; that work forms the experimental baseline against which the present system is compared.

\begin{figure}[tbp]
    \centering
    \includegraphics[width=0.45\textwidth]{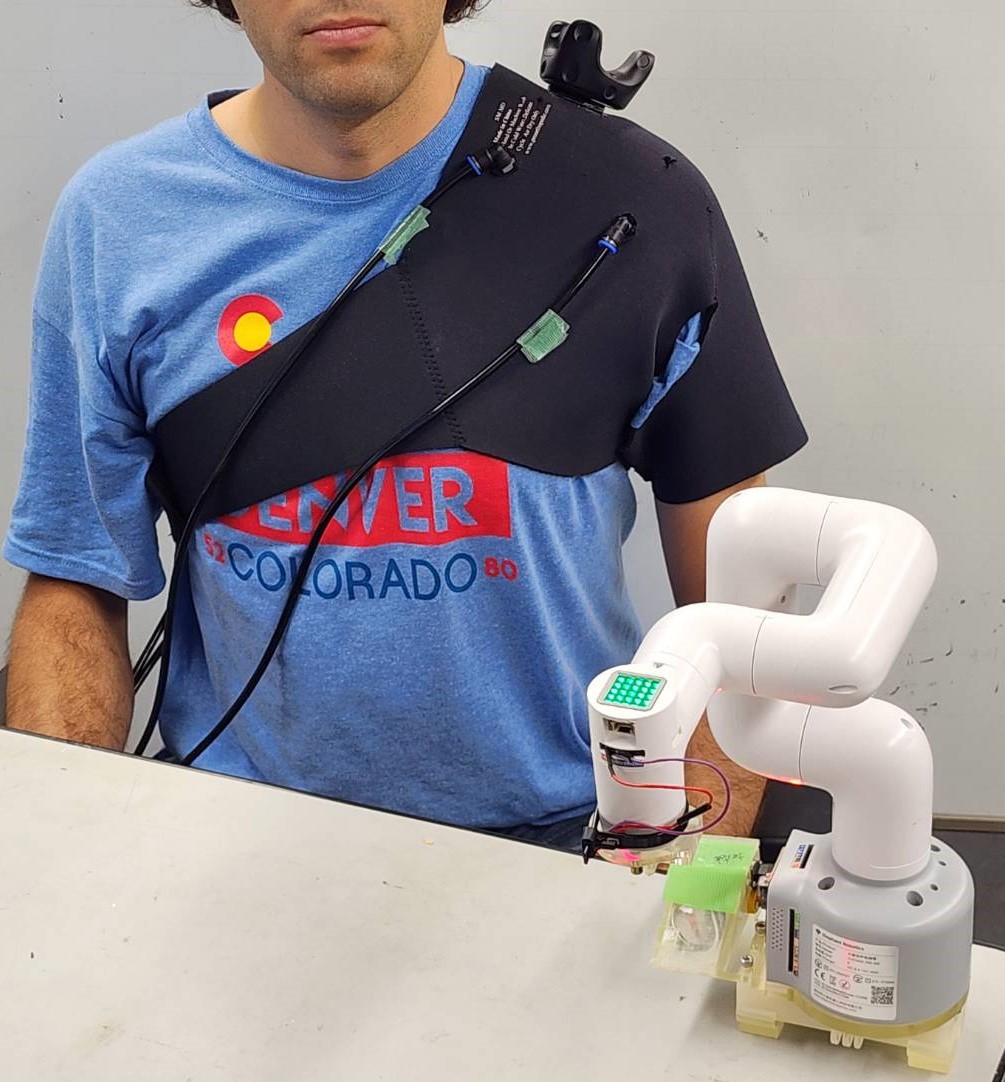}
    \caption{Illustration of MESSPa in use with the arm detached (experimental placement: Fig.~\ref{fig:expsetup}a). The HTC VIVE tracker on top of the shoulder pad maps the operator's shoulder motion to the arm one-to-one, while four shoulder-mounted balloons deliver positional haptic feedback. The remote arm and the task screen are visible through the operator's view.}
    \label{fig:MESSPA}
\end{figure}

Despite these developments, a central challenge remains: maintaining a high sense of agency and ownership across different states of wear. When the arm is attached, operators typically report a strong sense of control; when it is detached, this sense of agency---and, with it, performance---tends to diminish. This motivates the two research questions of this study:
\begin{itemize}
    \item Can a consistent sense of agency be maintained whether the robotic arm is attached or detached?
    \item Can the sense of ownership be enhanced to further improve task performance?
\end{itemize}

To address these questions we propose the Multi-modal Embodiment Stimulation Shoulder Pad (MESSPa, Fig.~\ref{fig:MESSPA}), which integrates two mechanisms: (i) one-to-one motion transfer based on an HTC VIVE tracker, so that the operator drives the arm with their own body even when it is detached, and (ii) spatially distributed pneumatic (balloon) haptic feedback on the shoulder that signals the arm's position. MESSPa is \emph{multi-modal} in that it adds two sensory channels to the operator's view of the arm: kinesthetic, through motion transfer that makes the operator's own movement drive the arm, and tactile, through pressure cues that signal the arm's position \cite{Ownership_FB_BALLOON}. The contributions of this paper are:
\begin{enumerate}
    \item the design of a shoulder-pad interface that combines motion transfer with positional balloon haptics in a single wearable form factor (Section~\ref{sec:system});
    \item a controlled $2\times2$ mixed-design evaluation crossing the control interface (MESSPa, $N=13$, vs.\ keyboard, $N=12$) and attachment under two cognitive-load levels (Section~\ref{sec:methods}); and
    \item empirical evidence that the interface preserves the sense of agency and improves task performance in the detached condition, narrowing the gap to the attached condition (Sections~\ref{sec:results}--\ref{sec:discussion}).
\end{enumerate}
We emphasize that balloon haptics is not itself new (Section~\ref{sec:relwork}); the contribution lies in applying spatially distributed shoulder haptics to a \emph{detachable} SRL to preserve agency during remote operation.

\section{Related Work}\label{sec:relwork}

\subsection{Wearable Robots and Supernumerary Robotic Limbs}
SRLs add extra, independently actuated limbs to the human body to augment manipulation and support \cite{Bright2017SupernumeraryRL}. Examples include waist- and shoulder-mounted extra arms for overhead assembly and load support \cite{Parietti2016_SRL, Llorens-Bonilla2012}, extra fingers for grasping \cite{Prattichizzo_extrafinger}, and exoskeletons that assist locomotion and rehabilitation \cite{Siviy2023_Exoskeleton, Sasaki2017}. A recurring finding in this literature is that the value of an SRL depends not only on its mechanical capability but on how well the operator can control it intuitively---that is, on the quality of embodiment it affords \cite{Inoue2019_Enhancing_bodily}.

\subsection{Detachable Bodies and Remote Operation}
To overcome the spatial constraint of body-worn limbs, modular and detachable robots allow the operator to leave the limb at a remote site and act in two locations at once \cite{Iwasaki2020, Iwasaki2021}. Fusion-style wearable teleoperation \cite{Fusion_Saraiji2008} and detachable end-effectors \cite{epfl2024detachable} pursue similar goals, while ubiquitous robotics frames these systems as part of a networked ecosystem \cite{Ubiquitous_robotics}. A key difficulty in remote operation is that separating the limb from the body weakens the perceptual coupling between intention and action and can be aggravated by latency \cite{impact_of_latency}. Our earlier work \cite{Vimolmongkolporn2023} characterized this degradation for a modular head--arm body and established the attached/detached baseline used here.

Telexistence systems pursue embodiment through dedicated master devices: TELESAR~VI \cite{Tachi2020} combines a full-body cockpit, an anthropomorphic avatar, and fingertip haptics, and reviews discuss how embodiment can enhance teleoperation and how to design for and assess agency and ownership over a remote robot \cite{Toet2020, Falcone2022}. MESSPa instead targets a limb that alternates between attached and detached use: it keeps the same body-anchored mapping in both states and adds positional cues at the shoulder, where the arm is mounted when attached.

\subsection{Sense of Embodiment: Agency and Ownership}
The sense of embodiment is widely modeled in terms of agency and ownership \cite{AgencyandOwnership, BLANKE20097_SOE}. Ownership has been studied through the rubber-hand illusion \cite{Botvinick1998} and extended to robotic and supernumerary limbs \cite{Kieliba2021_robotichand}, whereas agency is closely tied to the temporal and spatial congruence between a user's action and its sensory consequences \cite{Haggard2012_SenseofAgency, Kondo2018_bodyOwnership}. We quantify both constructs with the standardized embodiment questionnaire of Gonzalez-Franco and Peck \cite{Gonzalez-Franco2018}, adapted to the present task.

\subsection{Haptic Feedback in Teleoperation}
Haptic feedback improves control and embodiment in teleoperation, but conventional vibrotactile cues can become uncomfortable or fade through desensitization after prolonged actuation \cite{hapticFB_SURVEY, Toet2020}. Pneumatic balloon actuators provide a softer, pressure-based alternative and have been used to convey contact and to enhance ownership \cite{Hapballoon, Ownership_FB_BALLOON}. We adopt balloon actuation as an established technique; our novelty is its spatial arrangement on the shoulder to encode the remote arm's position and thereby sustain the sense of agency in the detached state.

\subsection{Dual-Task Load}
Real deployments require operating the arm while attending to a concurrent task. Performing two tasks at once introduces interference and increases cognitive load \cite{Pashler1994, Monsell2003, Sweller1988}. We therefore evaluate MESSPa under both single- and dual-task conditions so that its effect can be assessed under realistic load.

\section{System Design}\label{sec:system}
\begin{figure}[tp]
    \centering
    \includegraphics[width=\columnwidth]{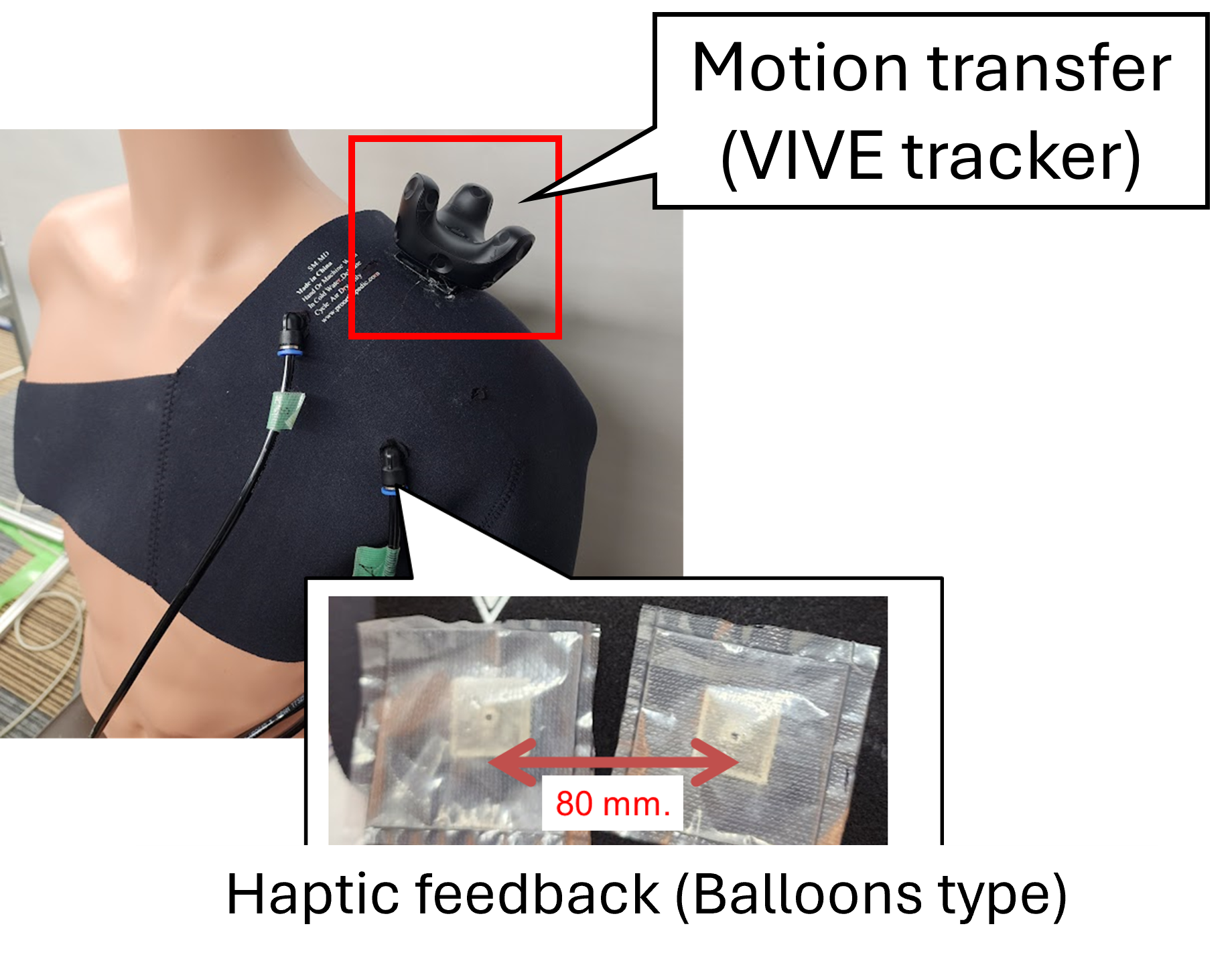}
    \caption{The MESSPa shoulder pad. Motion transfer is provided by an HTC VIVE tracker; haptic feedback is provided by four balloons (25$\times$25~mm each) arranged on the shoulder with 80~mm spacing, corresponding to the four quadrants of the arm's workspace (Section~\ref{subsec:haptic}).}
    \label{fig:hapticsuit}
\end{figure}

This section describes the MESSPa prototype, whose two functional components---motion transfer and haptic feedback---are designed to support a seamless and intuitive operating experience (Fig.~\ref{fig:hapticsuit}).

\subsection{System Requirements}
From the goal of preserving agency across wear states, we identify two requirements: (i) \emph{motion transfer}, so the operator drives the arm with their own body and retains a direct action--consequence coupling even when the arm is detached; and (ii) \emph{positional haptic feedback}, so the operator perceives where the remote arm is without relying solely on vision.

\subsection{Motion Transfer}\label{subsec:motion}
To support agency, the operator's shoulder and upper-body motion is captured by a single HTC VIVE tracker \cite{vive_tracker_3_0} mounted on top of the shoulder pad (Fig.~\ref{fig:hapticsuit}) and mapped one-to-one onto the robotic arm; no tracker or controller is worn on or held in the hand. We use a \emph{relative} pose mapping: at the start of a trial the tracker pose is recorded as the reference $\vec{r}_{0}$, and the command sent to the arm is the displacement of the current tracker pose $\vec{r}_{tracker}$ from this reference,
\begin{equation}
\vec{r}_{robot} = \vec{r}_{tracker} - \vec{r}_{0} =
\begin{bmatrix}
r_{X} - r_{X,0} \\
r_{Y} - r_{Y,0} \\
r_{Z} - r_{Z,0} \\
r_{rotX} - r_{rotX,0} \\
r_{rotY} - r_{rotY,0} \\
r_{rotZ} - r_{rotZ,0} \\
0
\end{bmatrix},
\label{eq:handControl}
\end{equation}
where the first three entries are the translation and the next three the orientation of the tracker relative to the reference, both expressed in the tracker frame and applied to the arm's base frame. A relative (rather than absolute) mapping lets the operator re-centre comfortably and makes control independent of where the tracker happens to start. The seventh entry is fixed to zero: it corresponds to the gripper degree of freedom, which is not actuated in this task. For the experiment the mapping is restricted to the two horizontal Cartesian axes (a 2-DoF planar mapping) to match the planar pointing task. The resulting command is streamed from Unity \cite{unityengine} to a MyCobot~280 \cite{elephant_robotics_mycobot_2024} through the \texttt{pymycobot} API \cite{pymycobot_2024}.

\subsection{Haptic Feedback}\label{subsec:haptic}
Conventional vibrotactile feedback, while effective, can become uncomfortable or lose salience after extended actuation \cite{hapticFB_SURVEY}. We therefore use pneumatic balloon feedback \cite{Hapballoon, Ownership_FB_BALLOON}, in which air-filled compartments expand and contract in response to the operator's interaction with the arm, providing a softer and more sustained tactile cue.

\subsubsection{Placement and Number}
The balloons are placed on the shoulder, close to the natural point of attachment and control of the arm, so that feedback is immediate and spatially relevant. Four balloons are used, one per quadrant of the arm's planar workspace around its starting position, giving coverage of the shoulder area sufficient to indicate direction.

\subsubsection{Size and Spacing}
Each balloon measures 25$\times$25~mm---large enough to produce a clearly perceptible cue but small enough to remain unobtrusive under a garment. The centre-to-centre spacing is 80~mm (Fig.~\ref{fig:hapticsuit}). This value is chosen with reference to the two-point discrimination threshold on the upper body, reported in the range 31--44~mm \cite{nolan1982twopoint}; an 80~mm spacing comfortably exceeds this threshold, so that each balloon is felt as a distinct stimulus rather than blurring into its neighbours.

\subsection{Pneumatic System}
\begin{figure}[tbp]
    \centering
    \includegraphics[width=0.45\textwidth]{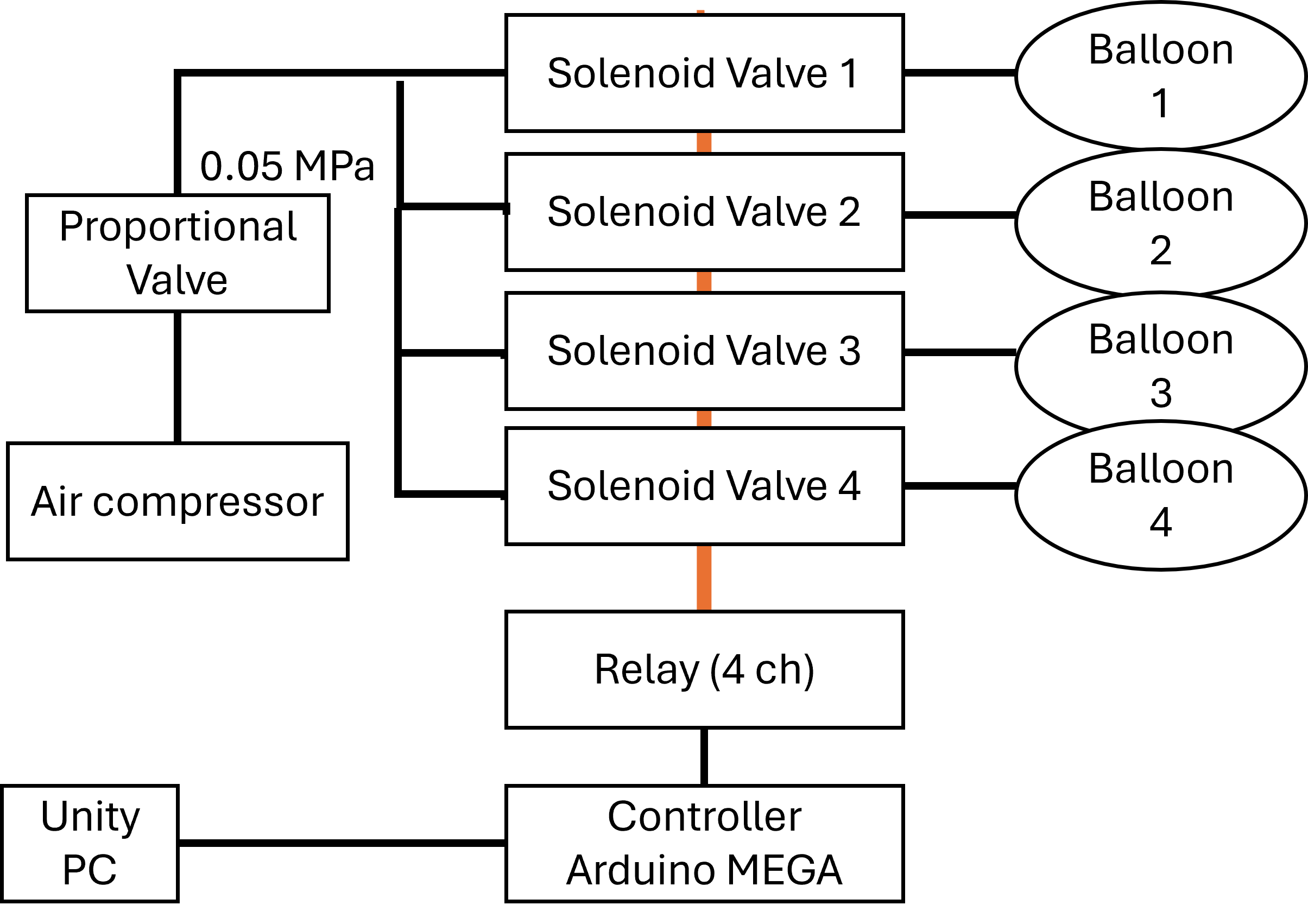}
    \caption{Pneumatic system. The compressor supply is regulated to 0.05~MPa by a proportional valve; four solenoid valves driven by an Arduino MEGA inflate the balloons according to the arm's quadrant, as commanded by Unity.}
    \label{fig:Pneumatic}
\end{figure}

The pneumatic system (Fig.~\ref{fig:Pneumatic}) supplies and regulates air pressure to the balloons. A compressor generates the supply pressure, which a proportional valve \cite{p_valve} regulates to 0.05~MPa---the value established in our previous study \cite{Vimolmongkolporn2023}. Solenoid valves (030E1-PSL DC24V) \cite{SolenodValve} connected to each balloon are driven by an Arduino MEGA \cite{arduino_mega} that interfaces with Unity. Unity determines which of the four workspace quadrants the arm's end-effector currently occupies and activates the corresponding balloon.

\FloatBarrier

\begin{figure}[htbp]
    \centering
    \includegraphics[width=\columnwidth]{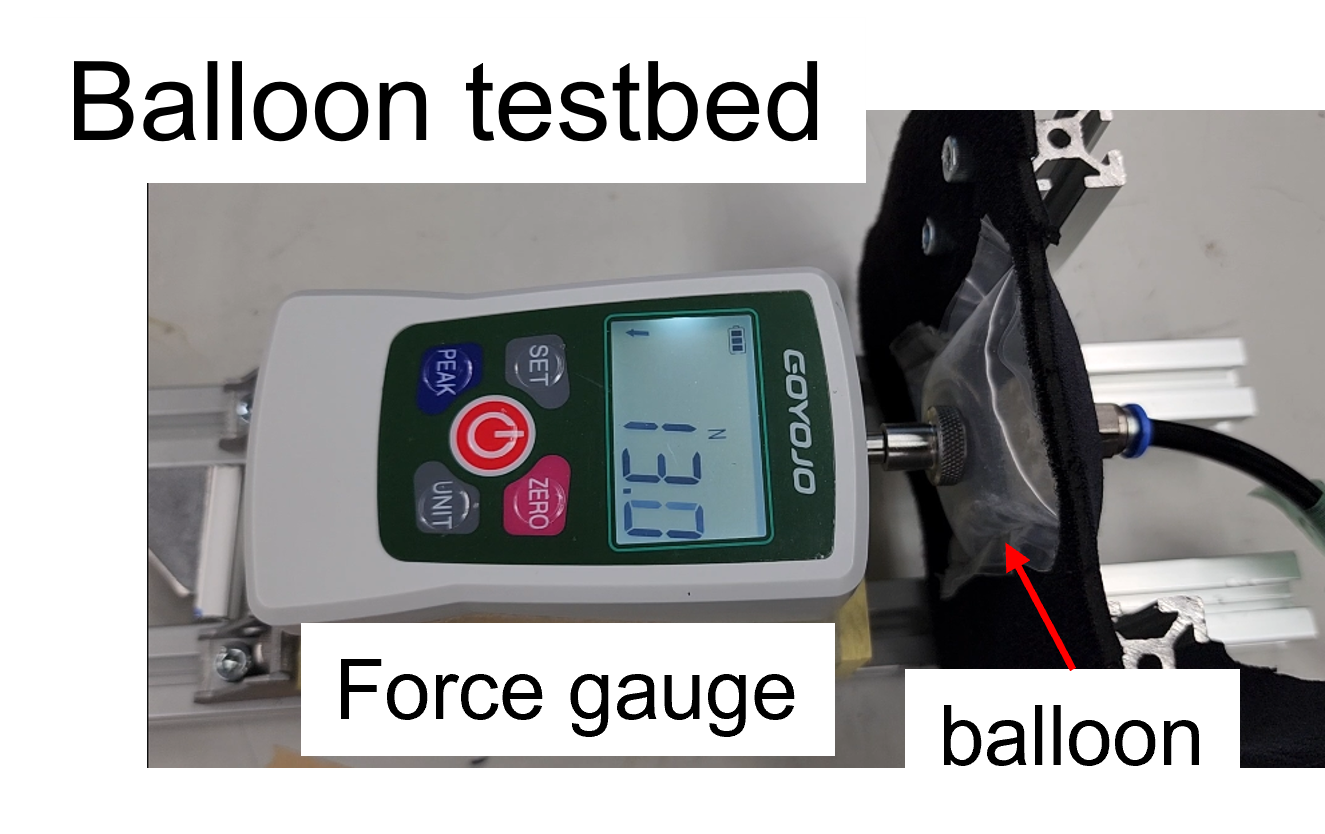}
    \caption{Balloon test bed used for force validation, consisting of a force gauge and a balloon locking mechanism that holds the balloon against the gauge while the input pressure is varied.}
    \label{fig:Ch4_BalloonTestBed}
\end{figure}

\subsubsection{Force Validation}
To confirm the force delivered to the skin, we measured the force at the centre of a balloon on a test bed with a force gauge (Fig.~\ref{fig:Ch4_BalloonTestBed}), which provides repeatable, well-localized measurements. At the operating pressure of 0.05~MPa \cite{Vimolmongkolporn2023}, a balloon exerts a maximum force of 13~N. This level was selected to be clearly perceptible without causing discomfort: it provides a salient cue while remaining below the threshold at which participants reported any discomfort.
\section{Methods}\label{sec:methods}
We evaluated the effect of MESSPa on task performance and on the sense of embodiment in a controlled experiment.

\subsection{Experimental Design}\label{subsec:design}
\begin{figure*}[htbp]
    \centering
    \includegraphics[width=\textwidth]{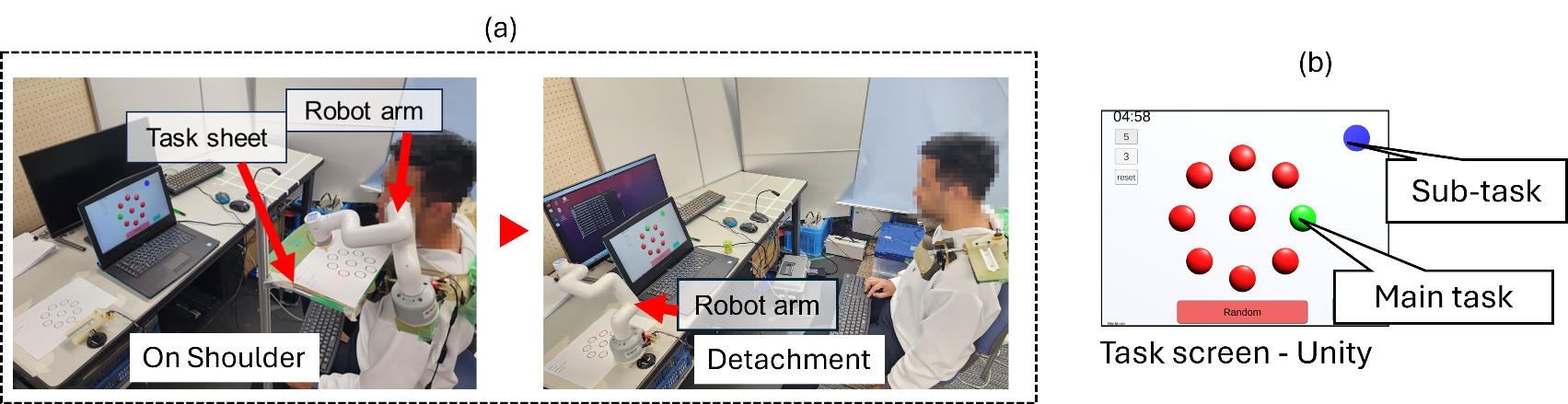}
    \caption{(a) Experimental setup, showing the on-shoulder (attached) and detached conditions without MESSPa. (b) The Unity task screen, with the timer, the main laser-pointing target (green circle), and the sub-task colour indicator.}
    \label{fig:expsetup}
\end{figure*}

The experiment used a $2\times2$ mixed design crossing two factors: \emph{interface} (with MESSPa vs.\ without MESSPa, in which the arm was controlled with a keyboard), varied between groups, and \emph{attachment} (on-shoulder vs.\ detached), varied within participants. The MESSPa group comprised 13 participants; the keyboard data are those of the 12 participants of our previous study \cite{Vimolmongkolporn2023}. This yields the four wearing conditions analysed in Section~\ref{sec:results}: on-shoulder/without, detached/without, on-shoulder/with, and detached/with MESSPa. Each factor combination was performed under two cognitive-load levels, a single task and a dual task (Section~\ref{subsec:tasks}). In the MESSPa group, the order of the attachment conditions and of the single and dual tasks was counterbalanced; the keyboard group followed the procedure of \cite{Vimolmongkolporn2023}. The experimental setup is shown in Fig.~\ref{fig:expsetup}.

\subsection{Participants}
Thirteen volunteers took part in the MESSPa group (12 male, 1 female; age 21--30 years, $M=24.7$, $SD=2.59$). The keyboard group consisted of the 12 participants of our previous study~\cite{Vimolmongkolporn2023} (11 male, 1 female; age 21--30 years, $M=24.3$, $SD=2.34$). All were students or laboratory members of the university and participated voluntarily after giving informed consent.

This study met the criteria for exemption from ethics review under the Waseda University guideline for research involving human subjects: (i) only de-identified information was used; (ii) all participants were students or laboratory members who took part voluntarily with informed consent; and (iii) the methodology involved no invasive procedure and no direct contact between the body and the robot---interaction was mediated by a harness with a soft foam sheet, and no participant reported discomfort. The arm weighed 850~g and the harness 3.5~kg.

\subsection{Apparatus and Task}\label{subsec:tasks}
The task software was developed in Unity3D and displayed a timer, a main-task area, and a sub-task area (Fig.~\ref{fig:expsetup}b). In the on-shoulder condition the arm is mounted on a harness; with MESSPa, the shoulder pad was worn under the harness, with the balloons and the tracker clear of the harness straps so that they were not compressed.

\paragraph{Main task --- laser pointing}
A green target circle appears at the centre of the screen. The \emph{task sheet} (Fig.~\ref{fig:expsetup}a) is a paper sheet printed with circles matching the on-screen layout: the green circle on the screen indicates which printed circle to aim at, and the laser dot lands on the sheet, where the external camera tracks it. The participant moves a laser pointer mounted on the arm onto the target and presses ``Enter'' on a small keypad. An external camera, processed in OpenCV \cite{opencv_library} by colour-thresholding and centroiding the laser dot, records the pointer position (converted from pixels to millimetres by a one-time calibration), and the time stamp is logged to a CSV file. The target then resets to the centre and the cycle repeats for five minutes.

\paragraph{Sub-task --- colour calling}
In the dual-task condition a coloured circle (red, blue, or green) is added to the screen and changes colour at random every 5--15~s; the participant must say the current colour aloud while performing the main task. This concurrent demand raises cognitive load \cite{Pashler1994, Sweller1988} and reflects the practical case in which the arm is operated while attending to something else. (A more demanding subtraction variant of the sub-task was used only in the long-term study discussed as future work in Section~\ref{sec:conclusion}, not in the present experiment.)

\subsection{Measures}
Performance was assessed objectively by \emph{task completion time} (seconds per target) and \emph{accuracy} (error distance in millimetres from the target centre). Embodiment was assessed subjectively with the standardized embodiment questionnaire \cite{Gonzalez-Franco2018}, reworded for this task (Table~\ref{tab:table1}) and restricted to agency and ownership. Following the questionnaire guideline, the component scores are computed as $\mathrm{SoA}=A1+A2+A3-A4$ and $\mathrm{SoO}=O1+O2$.

\begin{table}[htbp]
\centering
\caption{Sense of Embodiment Questionnaire (adapted from \cite{Gonzalez-Franco2018}).\label{tab:table1}}
\begin{tabular}{|p{0.02\textwidth}|p{0.37\textwidth}|}
\hline
\multicolumn{2}{|l|}{\textbf{Sense of Agency (SoA)}} \\
\hline
A1 & It felt like I could control the robot arm as if it were my own arm \\
A2 & The movements of the robot arm were caused by my actions \\
A3 & I felt as if the movements of the robot arm were influencing my own movements \\
A4 & I felt as if the robot arm was moving by itself \\
\hline
\multicolumn{2}{|l|}{\textbf{Sense of Ownership (SoO)}} \\
\hline
O1 & I felt as if the robot arm was a part of my body \\
O2 & I felt as if the robot arm belonged to me \\
\hline
\end{tabular}
\end{table}

\subsection{Procedure}
Each participant of the MESSPa group completed the two attachment conditions in three phases.
\begin{enumerate}
    \item \textbf{Practice.} Participants practised both the single and the dual task for at least three minutes each, in the detached configuration, to learn the one-to-one mapping (a 2-DoF horizontal Cartesian mapping of body motion to the arm).
    \item \textbf{Attached.} Participants performed the tasks with the arm on the shoulder, then completed the embodiment questionnaire.
    \item \textbf{Detached.} The experimenter detached the arm and placed it at a remote position, confirming that the participant could still see the task area clearly. Participants repeated the tasks and the questionnaire.
\end{enumerate}
The order of conditions and of single/dual tasks was counterbalanced to minimize sequence effects.

\subsection{Statistical Analysis}\label{subsec:stats}
Normality was checked with the Shapiro--Wilk test \cite{SHAPIRO1965}. Because the interface was varied between groups and attachment within participants, we tested the three planned comparisons directly: on-shoulder vs.\ detached within each group (paired tests) and detached with vs.\ without MESSPa between groups (unpaired tests). Single-task measures were analysed with paired $t$-tests and Welch's $t$-test, dual-task measures with Wilcoxon signed-rank tests \cite{Wilcoxon} and Mann--Whitney $U$ tests. We report Bonferroni-adjusted $p$-values (three comparisons), evaluated against $\alpha=0.05$. For the detached-condition embodiment comparison we used an independent-samples $t$-test. Analyses were performed in Python with SciPy~1.18.

\section{Results}\label{sec:results}
We report task completion time, accuracy, and the sense of embodiment. Each measure is analysed across the four wearing conditions defined in Section~\ref{subsec:design}.

\subsection{Task Completion Time}
\begin{figure}[tbp]
    \centering
    \includegraphics[width=\columnwidth]{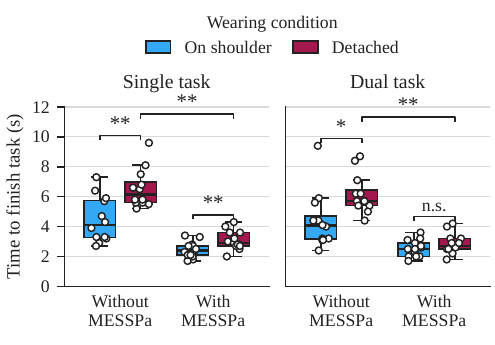}
    \caption{Task completion time across the four wearing conditions, for the single and dual task, with the baseline from the previous experiment \cite{Vimolmongkolporn2023}. Brackets mark the three planned comparisons: ** $p<0.01$, * $p<0.05$ (Bonferroni-adjusted), n.s.\ not significant.}
    \label{fig:Time_comparisonFrompvexp}
\end{figure}

Fig.~\ref{fig:Time_comparisonFrompvexp} compares completion time with and without MESSPa, including the baseline from the previous experiment \cite{Vimolmongkolporn2023}.

For the single task: (i) on-shoulder was significantly faster than detached without MESSPa ($t(11)=-6.888$, $p<0.001$); (ii) on-shoulder was also significantly faster than detached \emph{with} MESSPa ($t(12)=-4.012$, $p=0.005$); and (iii) detached \emph{with} MESSPa was significantly faster than detached without MESSPa (Welch $t(15.7)=8.365$, $p<0.001$).

For the dual task: (i) on-shoulder was faster than detached without MESSPa ($Z=-2.845$, $p=0.013$); (ii) on-shoulder vs.\ detached with MESSPa did not differ significantly ($Z=-2.205$, $p=0.082$); and (iii) detached with MESSPa was faster than detached without MESSPa ($U=0$, $p<0.001$).

\subsection{Accuracy (Error Distance)}
\begin{figure}[tbp]
    \centering
    \includegraphics[width=\columnwidth]{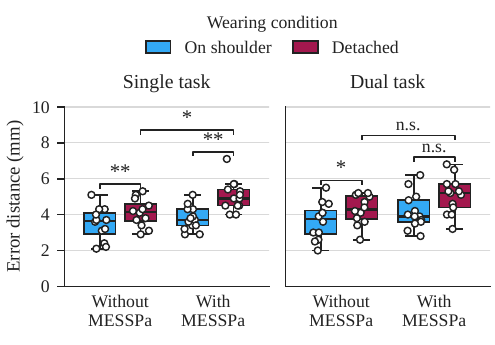}
    \caption{Error distance (mm) across the four wearing conditions, for the single and dual task. Brackets mark the three planned comparisons: ** $p<0.01$, * $p<0.05$ (Bonferroni-adjusted), n.s.\ not significant.}
    \label{fig:Accuracy_comparisonFrompvexp}
\end{figure}

Fig.~\ref{fig:Accuracy_comparisonFrompvexp} compares accuracy with and without MESSPa.

For the single task: (i) on-shoulder was significantly more accurate than detached without MESSPa ($t(11)=-4.439$, $p=0.003$); (ii) on-shoulder was also significantly more accurate than detached with MESSPa ($t(12)=-5.487$, $p<0.001$); and (iii) detached \emph{without} MESSPa was significantly more accurate than detached \emph{with} MESSPa (Welch $t(23.0)=-2.925$, $p=0.023$).

For the dual task: (i) on-shoulder was more accurate than detached without MESSPa ($Z=-2.630$, $p=0.026$); (ii) on-shoulder vs.\ detached with MESSPa did not differ significantly ($Z=-2.271$, $p=0.069$); and (iii) detached with vs.\ without MESSPa did not differ ($U=41$, $p=0.140$).

\subsection{Sense of Ownership and Agency}
\begin{figure}[tbp]
    \centering
    \includegraphics[width=0.45\textwidth]{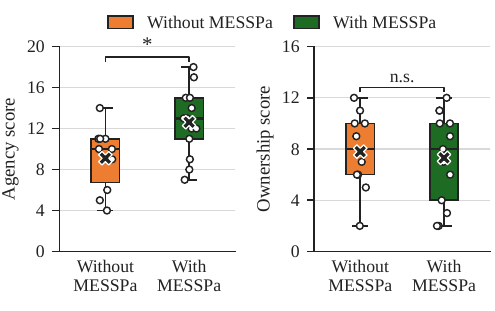}
    \caption{Sense of agency and sense of ownership in the detached condition, with and without MESSPa. Agency improves significantly with MESSPa ($p=0.017$); ownership is unchanged.}
    \label{fig:SoE_comparisonFrompvexp}
\end{figure}

We compared SoA and SoO with and without MESSPa in the detached condition using independent-samples $t$-tests (Fig.~\ref{fig:SoE_comparisonFrompvexp}). Agency improved significantly with MESSPa ($t=2.582$, $df=21$, $p=0.017$). Ownership showed no significant change ($t=-0.345$, $df=21$, $p=0.734$).

\section{Discussion}\label{sec:discussion}

\subsection{Task Completion Time and Accuracy}
The completion-time results show a clear benefit of MESSPa in the detached state. With keyboard control (without MESSPa), detaching the arm slowed participants significantly relative to the on-shoulder condition, in both single and dual tasks. With MESSPa, detached operation was significantly faster than with the keyboard, and the cost of detachment was smaller (mean increase of 0.6~s vs.\ 2.1~s per target in the single task, 0.3~s vs.\ 1.8~s in the dual task); it remained significant in the single task but not in the dual task. In other words, MESSPa narrowed, but did not fully close, the performance gap that detachment opens, which is consistent with the goal of preserving agency across wear states.

The accuracy results are more nuanced. In the single task, accuracy was actually \emph{lower} with MESSPa in the detached condition. This may reflect the balloon stimulus drawing attention away from fine positioning, or the difference between body-motion and keyboard control, when the cognitive load is low and time pressure is the dominant demand. Under the dual task, however, the accuracy difference between interfaces was no longer significant, and with MESSPa the difference between on-shoulder and detached operation was not significant either, whereas it was with the keyboard. This indicates that under realistic, higher load the accuracy cost of MESSPa disappears while its benefit to speed and consistency is retained. We note that the absolute error distances are small (a few millimetres), so the task may approach a precision ceiling; future work should use a more demanding positioning task to separate these effects.

Detaching the arm also changes the viewing distance and angle (Fig.~\ref{fig:expsetup}a), so the on-shoulder versus detached contrast compares two configurations rather than body contact alone. However, with MESSPa the completion-time cost of detachment was much smaller than with the keyboard, and not significant in the dual task, although the viewpoints differed in the same way; a purely visual account would predict similar costs for both interfaces. Between the two detached conditions, the arm was placed at the same angle and at nearly the same distance, so viewing geometry is unlikely to explain the effect of MESSPa.

\subsection{Sense of Embodiment}
MESSPa significantly improved the sense of agency in the detached condition, indicating that real-time motion transfer combined with positional haptic feedback strengthens the felt coupling between the operator's intention and the arm's action. The sense of ownership, by contrast, did not change significantly. A plausible explanation is that detachment imposes a large perceptual separation between body and limb that short, single-session feedback cannot overcome; building ownership likely requires repeated or prolonged exposure. This motivates the long-term study outlined below.

Unlike prior balloon feedback that enhanced ownership of a virtual hand by stimulating the hand itself \cite{Ownership_FB_BALLOON}, MESSPa delivers a coarse positional cue at the shoulder, which may support control more than body-part correspondence. Conversely, completion time improved alongside agency while ownership was unchanged, suggesting that securing agency is sufficient to recover most of the detachment penalty reported previously \cite{Vimolmongkolporn2023}, consistent with action--outcome accounts of agency \cite{Haggard2012_SenseofAgency}; this link is correlational, and ownership may still matter for longer or finer tasks.

\subsection{Limitations}
The samples were modest ($N=13$ with MESSPa, $N=12$ with the keyboard) and skewed toward male participants, and the task, though sensitive to speed effects, may be too easy to reveal accuracy differences. Attachment was also confounded with viewing geometry. The interface was compared between groups, and the keyboard data come from our previous study, so differences between groups and sessions cannot be excluded. Because MESSPa was compared with keyboard control, the separate contributions of motion transfer and of haptic feedback cannot be isolated. The study evaluated a single feedback encoding (quadrant-based inflation); richer spatial or intensity encodings remain to be explored.

\section{Conclusion and Future Work}\label{sec:conclusion}
We presented MESSPa, a shoulder-pad interface that combines one-to-one motion transfer with spatially distributed balloon haptics to preserve the sense of agency when a robotic arm is detached from the body. In a $2\times2$ mixed-design experiment ($N=13$ with MESSPa, $N=12$ with the keyboard), MESSPa significantly reduced task completion time in the detached condition---narrowing the gap to on-shoulder operation---and significantly improved the sense of agency, while leaving the sense of ownership unchanged.

Building on the ownership result, we are extending the evaluation to long-term wear. In a preliminary protocol, a small number of participants wear the arm for one to two hours per day over several consecutive days, using a more demanding subtraction sub-task to raise mental demand and reveal learning. The aim is to test whether repeated exposure allows the sense of ownership---and detached-condition performance---to converge toward the attached condition over time. We also plan to include a no-device baseline throughout the long-term protocol and to enlarge the sample so that the long-term effects can be tested statistically. A condition with the arm detached but placed beside the shoulder will separate body contact from viewing geometry. These directions should clarify how sustained use shapes embodiment in modular, detachable robotic systems.

\section*{Acknowledgment}
This research was supported by Waseda University Global Robot Academia Institute, Waseda University Green Computing System Research Organization, Japan.

\printbibliography[
title={References}
]

\end{document}